# A Simple and Efficient Registration of 3D Point Cloud and Image Data for Indoor Mobile Mapping System

Hao Ma[1], Jingbin Liu[1, *], Keke Liu[1], Hongyu Qiu[1], Dong Xu[1], Zemin Wang[1], Xiaodong Gong[1], Sheng Yang[1]

[1]State Key Laboratory of Information Engineering in Survering, Mapping and Remote Sensing, Wuhan University, Wuhan 430079, China - {mahao_fido, jingbin.liu, kekeliu, qiuhongyu, Whusggxudong, zeminwang, gongxiaodong, shengy}@whu.edu.cn

**Abstract:** Registration of 3D LiDAR point clouds with optical images is critical in the combination of multi-source data. Geometric misalignment originally exists in the pose data between LiDAR point clouds and optical images. To improve the accuracy of the initial pose and the applicability of the integration of 3D points and image data, we develop a simple but efficient registration method. We firstly extract point features from LiDAR point clouds and images: point features is extracted from single-frame LiDAR and point features from images using classical Canny method. Cost map is subsequently built based on Canny image edge detection. The optimization direction is guided by the cost map where low cost represents the the desired direction, and loss function is also considered to improve the robustness of the the purposed method. Experiments show pleasant results.

## 1. Introduction

Space digitization is the process of transforming scene information into three-dimensional digital products. It plays a very critical role in smart cities, Industry 4.0, artificial intelligence, security and transportation hub management, and has become a major domestic demand. The realization of spatial digitization usually requires research in the fields of computer vision, machine learning, and sensor fusion. Among them, the mobile mapping system (MMS) equipped with LiDAR and camera has become a very important digital method. Image and LiDAR point cloud data have their own advantages. Image can obtain the color information of the object, while the LiDAR point cloud data reflects the coordinate information of the object. Registration of LiDAR point cloud and image data can complement the two kinds of information to obtain a more complete and useful information. The registration between data obtained by different sensors, especially the registration of images and LiDAR, is the basis of data fusion content in geospatial information processing (Parmehr *et al.*, 2014). Normally, the camera and LiDAR are pre-calibrated, but the external orientation elements obtained by MMS are not accurate enough and cannot be directly used for registration of image sequences and moving laser scan data (Li *et al.*, 2018). This inaccuracy is mainly from the following aspects (Cui *et al.*, 2016): first, the influence of system errors. The pose measurement between different sensors on the mobile measurement system is not accurate enough; secondly, even the pose information provided by GPS/INS is usually not directly used for registration, and, in some cases, GPS/INS It cannot continuously provide the pose information of each frame. Especially when the existing parameters have large errors, or when the amount of data is relatively large, the automatic registration scheme becomes particularly important.

The registration of LiDAR and image data generally contains four steps (Cui *et al.*, 2016): registration primitives selection, similarity measure, registration transformation model and matching strategy. The first step of the direct registration algorithm is to extract features from images and point clouds. The features here are points, lines, and even semantic features, such as roofs, or stopped cars on the ground. After feature extraction or generation, a similarity measurement function is established to search for corresponding features. Different measurement methods may obtain different corresponding features. Therefore, it is extremely important to design a robust similarity measurement function, which inevitably influence the accuracy of the direct image registration parameters and the final registration effect. Although the registration of image and LiDAR data has attracted many researchers, there is no universal algorithm so far. On one hand, due to the development of airborne, vehicle-mounted, and indoor mobile measurement platforms, the data acquisition methods are diverse, and it is necessary to design a registration scheme for specific databases. On the other hand, because there are many technologies involved in the registration process, these existing technologies still need to be improved, especially in specific applications. So registration algorithm has a lot of room for improvement.

The registration problem of 2D image and 3D point cloud is converted into 2D to 2D registration. This type of method first generates an intensity image or a distance image from 3D point cloud, and then directly performs registration between the point cloud image and the image taken by the camera. Zhong et al. (Zhong *et al.*, 2009) extracted straight edges from the airborne LiDAR range image and aerial image respectively, and then obtained corner points and found corresponding feature points among these corners. Mutual information, as a measure of statistical similarity, has received extensive attention from scholars in the field of computer vision and remote sensing. Mutual information-based registration methods have been proposed in recent decades. Mastin et al. (Mastin *et al.*, 2009) used mutual information to calculate statistical dependence in data. Parmehr et al. (Parmehr *et al.*, 2014) adopted a joint mutual information technology to register images and LiDAR data, and also proposed a novel solution to the histogram size. Such methods have high computational complexity and poor real-time performance in the specific solution process. This idea of converting 2D to 3D registration into 2D to 2D registration uses the current mature image registration method. The disadvantage is that errors will be introduced when the image is generated from the point cloud, which subsequently affects the final registration accuracy. At the same time, this k ind of algorithm is not suitable for sparse LiDAR data.

The second type of algorithm is to convert the registration problem of 2D image and 3D point cloud into 3D to 3D registration. In order to overcome the problem of the large feature difference between the 3D point cloud and the 2D optical image, some scholars converted the registered data source, and used the principle of multi-view geometry to recover the 3D information from the image sequence, so that the registration of cloud and two-dimensional image was transformed into a spatial registration problem of two three-dimensional point clouds

(Zhang and Jiang, 2017). Leberl et al. (Leberl *et al.*, 2010) analyzed the point cloud generated from image sequence and LiDAR point cloud, and pointed out that there were many similarities between them. Compared with the direct registration of the image and the 3D point cloud mentioned above, registration between the two three-dimensional point sets is much easier to achieve. Stamos et al. (Stamos *et al.*, 2008) used the structure from motion (SfM) algorithm to recover the sparse 3D point cloud based on the SIFT feature extracted from image sequence, and finally registered the image point cloud with the LiDAR point cloud. Swart et al. (Swart *et al.*, 2011) proposed a non-rigid registration method based on the bundle adjustment integrated with iterative closest point algorithm (ICP), which aimed to solve the weak GPS signal. This algorithm is applicable for outdoor scenes where point features are relatively rich. Zheng et al.(Rongyong *et al.*, 2018) used a similar idea to first perform bundle adjustment on the image sequence, matched the resultant three-dimensional feature points with the LiDAR point cloud using the classic ICP algorithm, and deduced objective function according to the collinear equation. The nearest neighbor relationship in the ICP was put into the adjustment system, and the registration parameters were iteratively optimized. Bernard O.Abayowa et al. (Abayowa *et al.*, 2015) generated dense point clouds from optical aerial images using a reconstruction algorithm. The initial coarse parameters was computed from salient features extracted from the LiDAR and optical imagery-derived digital surface model (DSM) models, and the final registration parameters are further refined using the ICP algorithm to minimize global error between the registered point clouds. Hyojin Kim (Kim *et al.*, 2014) generated a disparity map from the stereo pair, and then registered the generated disparity map with that obtained by LiDAR. Based on the stereo image pairs, Fang Lv et al. (Lv and Ren, 2015) fused point features and line features to register optical images with LiDAR. Bisheng Yang first obtained the initial registration parameters through SfM and other technologies, and then utilized the ICP algorithm to register the dense point cloud generated by the stereo pair with the LiDAR point cloud (Yang and Chen, 2015). Chen Chi et al. (Chen *et al.*, 2018) proposed an automatic registration method for vehicle-mounted point clouds and panoramic images sequence. This method extracted the skyline vector from the 3D point cloud and the panoramic image respectively, and then used this line vector as the geometric registration primitive to solve the 2D-3D coarse registration model by minimizing the editing distance of the registration primitive. Finally, the ICP algorithm was used to perform 3D-3D registration between the dense point cloud generated by the panoramic image and the LiDAR to obtain the final registration parameters. The stereo pair-based method first generates a point cloud from images using the established relatively classic dense matching algorithm, and then performs the registration of the point cloud and the LiDAR point cloud to indirectly realize the registration of the image and the LiDAR point cloud. Most of the images in our experiments are captured in indoor corridors or relatively wide halls. Under this situation, the dense matching algorithm is not suitable because the ideal dense point cloud cannot be obtained. Therefore, the stereo matching method is not applicable.

The first two are indirect solutions, and there are also many studies that directly register optical images and LiDAR point clouds. Min Ding et al. (Ding *et al.*, 2008) used vanishing point and GPS information to estimate the initial external parameters of the camera, and then optimized the parameters using lines extracted from oblique aerial images and 3D model. This method is mainly suitable for aerial images, because the line in the indoor scene appears in the area where color is simple and single, so it is not ideal to rely on the corner points to extract the lines for indoor case. Lingyun Liu (Liu and Stamos, 2007) first adopted vanishing point information to obtain the rotation parameters of the camera, and then took the line feature as the registration primitive to obtain the object coordinates of the camera by minimizing the distance of the line feature with the same name. Lu Wang et al. (Wang and Neumann, 2009)) designed a block feature connected by edges according to the characteristics of the images taken in densely-built areas. Under the initial parameters, the features were extracted from the image and point cloud in advance and projection error was iteratively minimized to obtain the final registration parameters. This kind of method is suitable for outdoor scenes with buildings. According to the presence of stationary vehicles in urban scenes, Jiangping Li et al. (Li *et al.*, 2018) proposed a registration method that took vehicles as matching primitives. Although semantic features are used, this method also limits the scope of application of the algorithm. Zhang et al. (Zhang and Xiong, 2012) proposed a method to extract building contours directly from the airborne LiDAR point cloud, and then designed an automatic registration scheme using building corner features as the registration primitive. This method projected the LiDAR angle feature on the image based on the initial external orientation element of the aerial image, found the corresponding image angle feature according to a certain similarity measure, and finally used the intersection of two straight line segments in the angle feature as the control point for aerial imagery. The bundle adjustment was used to solve the external orientation elements. Lee et al. (Lee *et al.*, 2016) extracted the top information of the tree from the image and LiDAR point cloud acquired in the forest area, and used this feature to register the two data according to the collinear equation. As the author mentioned in their paper, this algorithm is only suitable for areas containing more trees, because it uses the top features of trees. Jung et al. (Jung *et al.*, 2016) designed a corner feature enhanced by the roof edge. Features were firstly extracted from aerial images and three-dimensional building models, and then based on the initial exterior orientation elements, the features extracted from the three-dimensional building object were projected to the image, the corresponding features were searched by the similarity measurement function, and then the registration parameters were calculated iteratively. This method is designed specially for the aerial image where roof feature is salient. Zhu et al. (Zhu *et al.*, 2018) proposed a registration method based on skyline features. They first extracted skyline from the image and LiDAR point cloud separately, and then solved registration parameters through the registration model derived from the collinear equation. Obviously, this method is suitable for ourdoor scenes where buildings are not particularly dense.

Most of the research is based on the registration of aerial images or outdoor images to LiDAR point clouds(Ravi *et al.*, 2018). There is little research on the registration of indoor images and 3D point clouds in the existing literature. However, according to the existing registration literature, it can be found that when studying the problem of image data and point cloud registration, it is crucial to extract appropriate and stable features for specific application scenarios. In this paper, we design a simple and efficient registration of indoor 3D point clouds and optical images. Point features are extracted from 3D point clouds frame by frame and image data using classical Canny edge detection. Cost map is further constructed to guide the optimization direction and loss function is introduced to improve the robustness of the optimization.

The remaining part of this paper is organized as follows. Point features are extracted from 3D

point Clouds and image data respectively. Then, registration parameters are optimized based on the designed cost map and the ceres library. Experiments are present in the subsequent section and conclusions are finally drawn.

## 2. Point Features Extraction

### 2.1. Point Features from 3D LiDAR points

We don't directly extract point features from 3D LiDAR point clouds, instead, single-frame LiDAR is considered here. Fig. 1(a) shows single-frame LiDAR points and Fig. 1(b) shows line detection result. Green points in Fig. 1(c) that represent start and end point will be removed. Other yellow points are detected as point features. To extract absolutely accurate point features is impractical. Although there are error point features among these yellow points, these features are rarely projected clear to that of image data as illustrated in section 2.1 and thus have little influence on the optimization progress. A example of point features extraction from a big data unit ( Fig. 2(a)) is shown in Fig. 2(b).

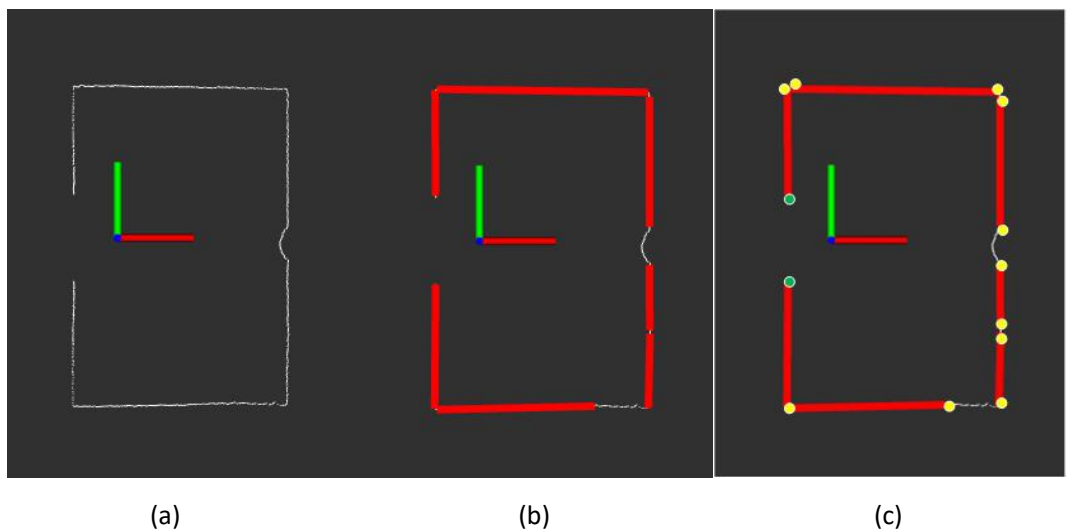

(a)                  (b)                  (c)

Fig. 1. Point features from single-frame LiDAR point clouds

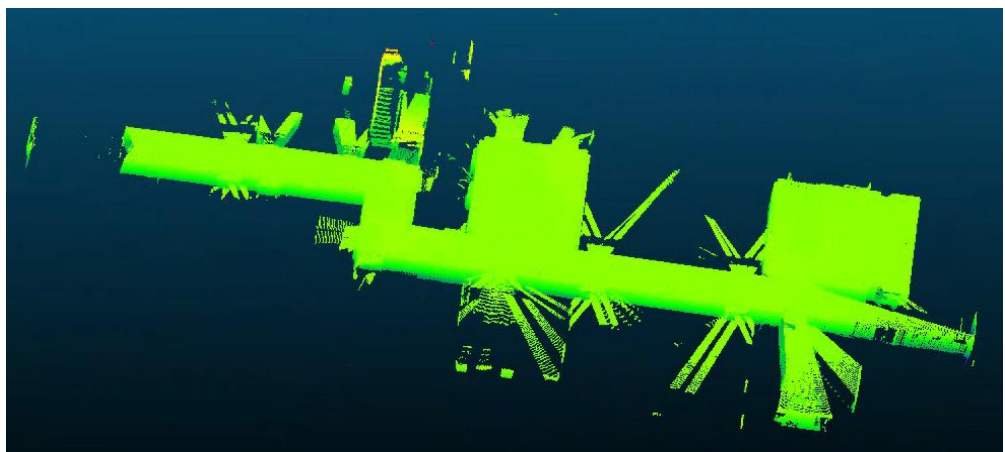

(a) 3D LiDAR point clouds

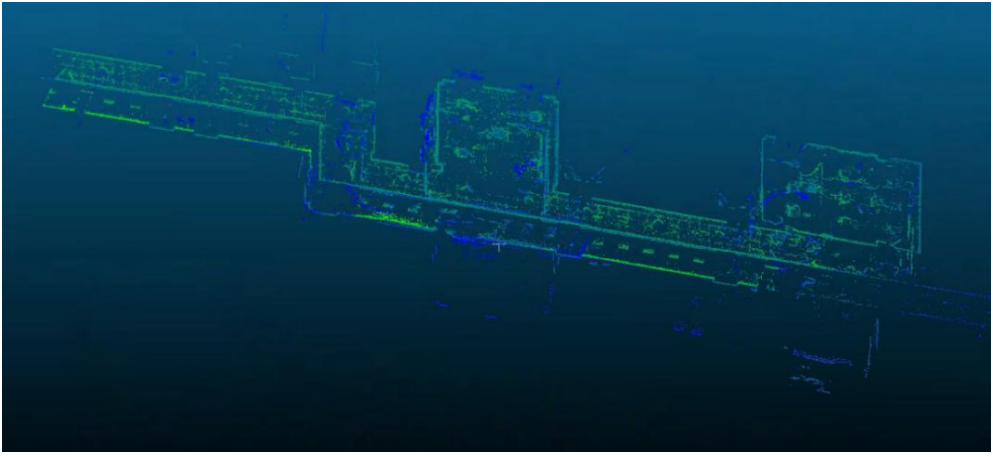

(b) detected point features

Fig. 2. Point features extraction from LiDAR point clouds

**2.2. Point Features from image data**

To extract point features from image data, the classical Canny edge detector is enough to meet our demand. To make it clear about Canny, we simply pick a concise description from related literature (Canny, 1986; Liu and Jezek, 2004). The Canny edge detector works in a multi-stage process. First, the image f(x,y) is smoothed by Gaussian convolution, then two-dimensional first derivatives are computed, the gradient magnitude and gradient direction are calculated. Suppose I(x,y) is the smoothed image. The first-order derivative of an image I(x,y) at location (x,y) is defined as the two-dimensional vector:

$$I_x(i,j) = \left(I(i,j+1) - I(i,j) + I(i+1,j+1) - I(i+1,j)\right)/2 \quad (1)$$

$$I_y(i,j) = \left(I(i,j+1) - I(i+1,j+1) + I(i,j) - I(i+1,j)\right)/2 \quad (2)$$

Gradient magnitude and gradient direction can be calculated through finite difference using first-order partial derivatives in 2×2 domain:

$$M(i,j) = \sqrt{I_x(i,j)^2 + I_y(i,j)^2} \quad (3)$$

$$\theta(i,j) = \arctan\left(\frac{I_y(i,j)}{I_x(i,j)}\right) - \frac{3}{4}\pi \quad (4)$$

Next, non-maximal suppression process is applied to the gradient magnitude (edge strength) image to identify the local maxima. Only pixels with edge strength larger than their two adjacent pixels in the gradient direction are identified as edge candidates and others are set to zero. Non-maximal suppression results in one-pixel wide edge segments. To remove false edge segments caused by noise and fine texture, a hysteresis tracking process is further applied with two thresholds in which all candidate edge pixels below the lower threshold are labelled as non-edges and all pixels above the low threshold that can be connected to any pixels above the high threshold through a chain of edge pixels are labelled as edge pixels.

There are three parameters to be set before using Canny: two thresholds and a aperture size. The ratio between the high threshold and the low threshold can be set in [2, 3], low threshold can be in [50, 100], and the aperture size should be a odd number between 3 to 7. We design a simple experiment for parameters analysis. Raw RGB image is shown in Fig. 3. Fig. 4 and Fig. 5 display edge detection results based different low thresholds and different aperture sizes respectively and black points represent detected point features. We can observe that lower threshold brings more point features, so does bigger aperture size. The principle of parameters selection in our registration is making as more point features that may have correspondences in that of 3D point features as possible and as less putative point features that do not have correspondences in that of 3D point features as possible. After a trade-off based on these experiments, low threshold, ratio and the aperture size are set as 50, 3, and 3 respectively. Readers who are interested in realizing our method and accept our advice on the setting of parameters can naturally obtain a satisfactory result.

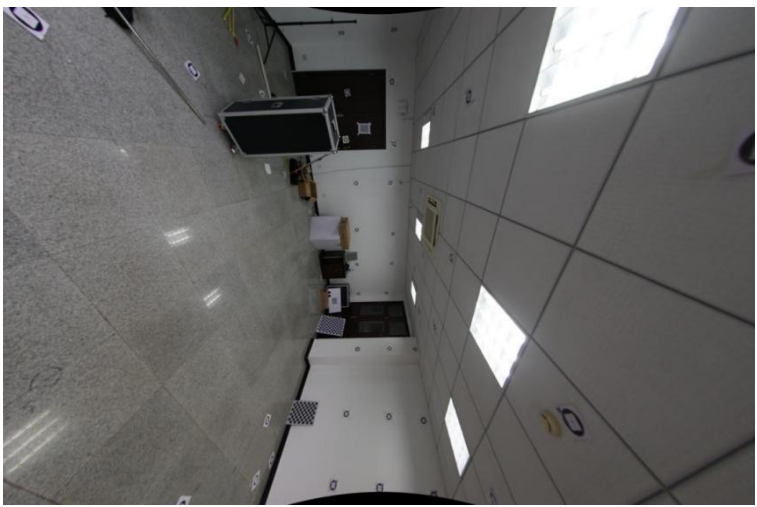

Fig. 3. Image data

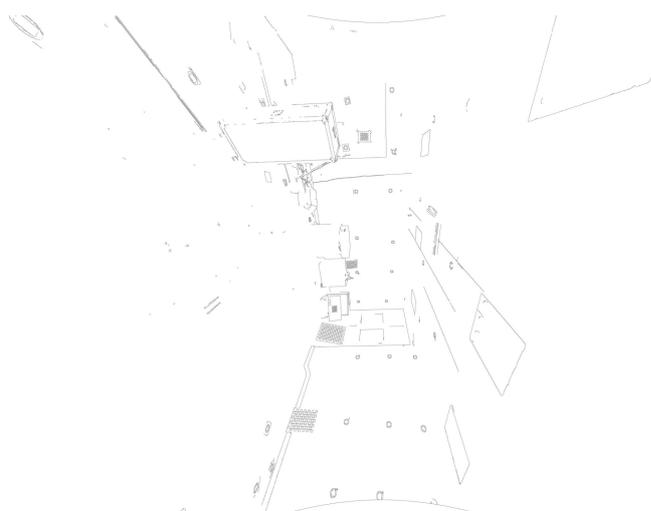

(a) high threshold/low threshold = 3, low threshold = 50, aperture size = 3

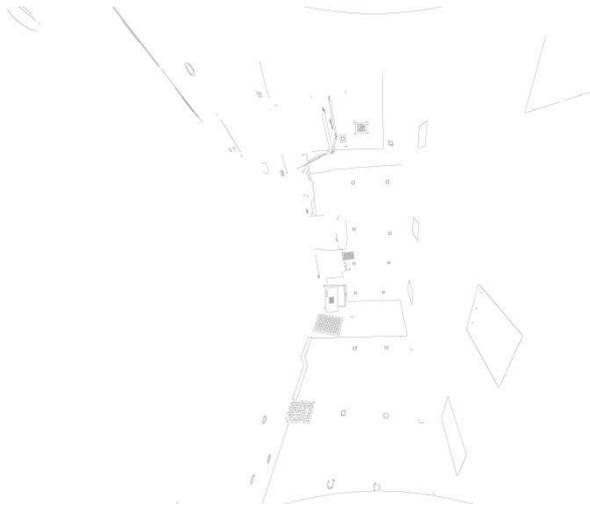

(b) high threshold/low threshold = 3, low threshold = 80, aperture size = 3

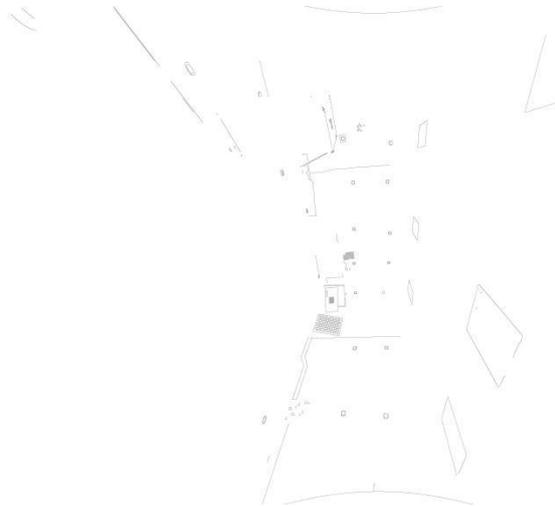

(c) high threshold/low threshold = 3, low threshold = 100, aperture size = 3

Fig. 4. Canny edge detection based on different low thresholds

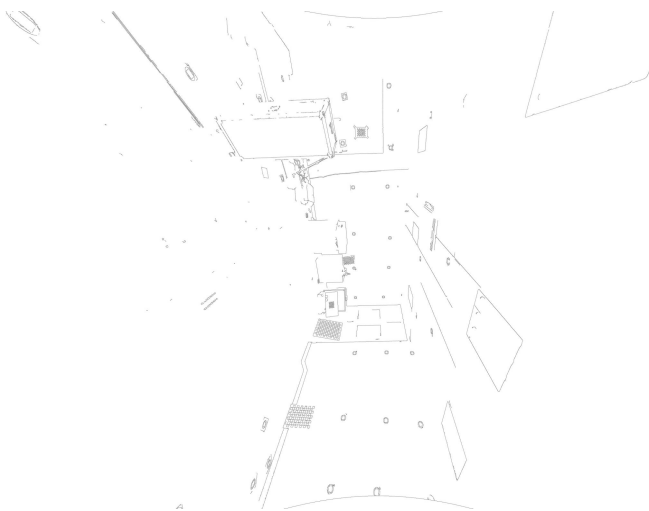

(a) high threshold/low threshold = 3, low threshold = 50, aperture size = 3

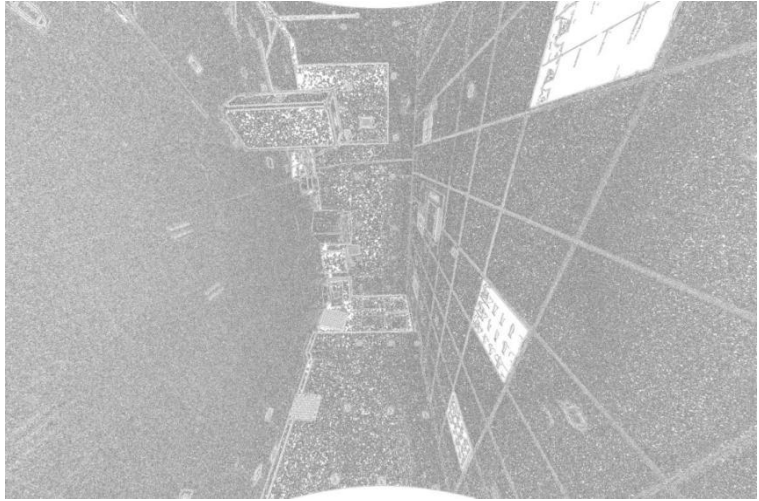

(b) high threshold/low threshold = 3, low threshold = 50, aperture size = 5

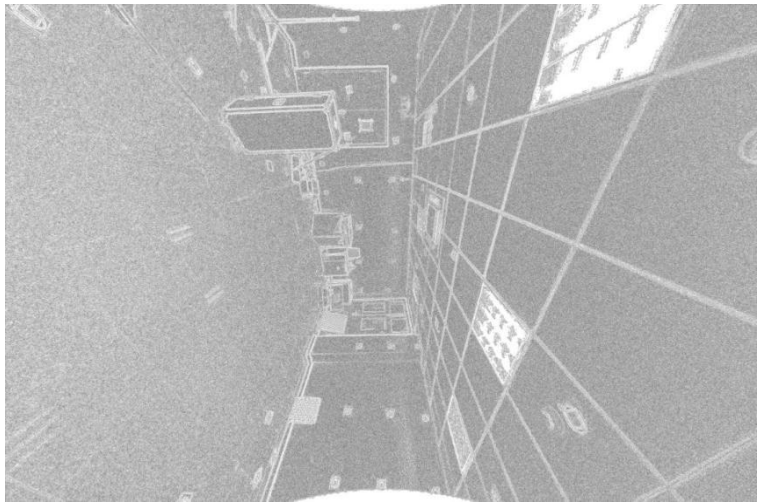

(c) high threshold/low threshold = 3, low threshold = 50, aperture size = 7

Fig. 5. Canny edge detection based on different aperture sizes

## 3.Point Features Based Optimization

### 3.1 depth based occlusion detection

Given initial registration parameters, we can project 3D point clouds onto image plane primarily. However, not all points projected onto the image plane will be taken into consideration during the registration progress. Fig. 6 shows different projection results without (Fig. 6(a)) and with (Fig. 6(b)) occlusion detection. Without occlusion detection, many point clouds that actually can not be captured by the camera are projected onto image plane. We need to deal with the situation where occlusion happens.

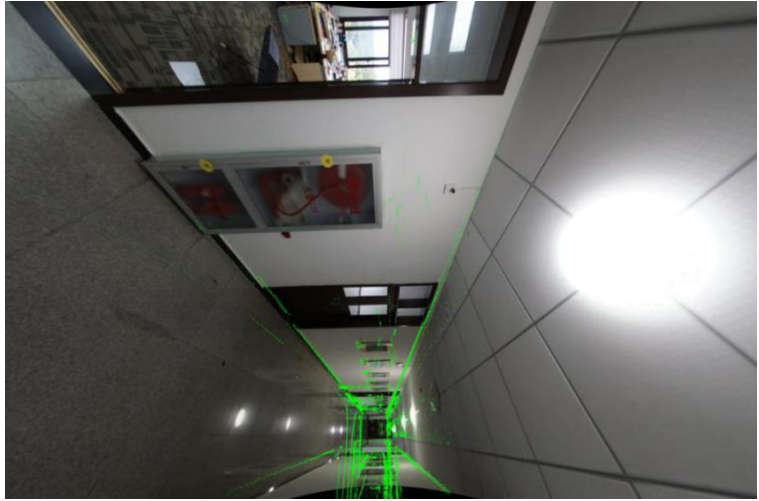
(a)

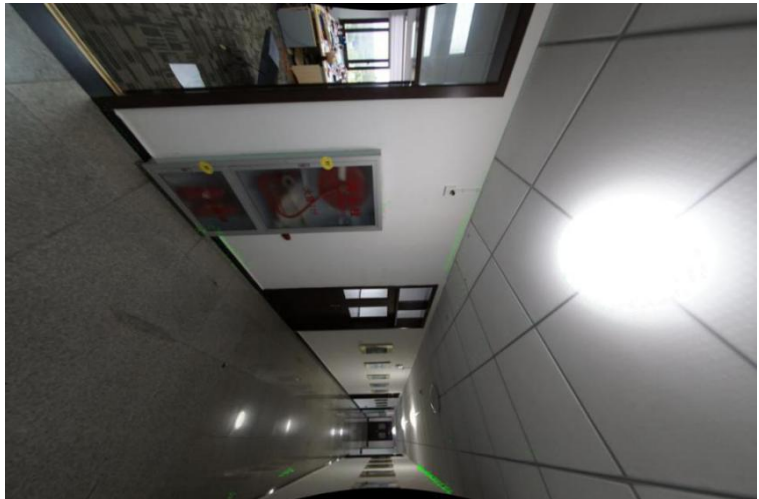
(b)

Fig. 6. Projection results without and with occlusion detection

In our paper, we utilize depth to carry out this issue. A depth map is generated ahead for a image. When a 3D point is projected onto the image plane and corresponds to (u, v), this 3D point will be abandoned if its depth value is bigger than that of (u, v) in the depth map.

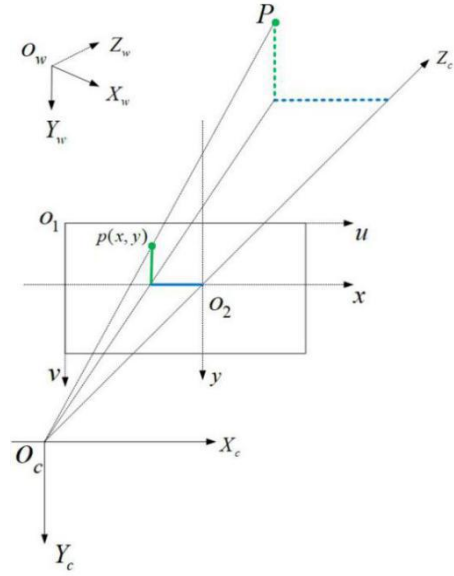

Fig. 7. Camera coordinate and world coordinate

To generate depth map, we need to make related transformations or coordinates clear. Transformation between 3D LiDAR scanner and a CCD camera involves several coordinates. Suppose the coordinate of 3D point cloud is $O_w$-$X_w Y_w Z_w$ and $O_c$-$X_c Y_c Z_c$ is the camera coordinate. $O_1$-uv is the pixel coordinate and $O_2$-xy is the image plane coordinate. p(x, y) on the image plane is the projection of P. Above mentioned coordiantes can be illustrated together in Fig. 7:
and their mathematical relationship is:

$$P_C = RP_W + t \tag{5}$$

where *R* and *t* are the rotation and translation between word and camera coordinate. Subscript in P denotes specific coordinate: 'C' means the camera coordinate and 'W' the world coordinate. The formulation above is equivalent to:

$$\begin{bmatrix} X_C \\ Y_C \\ Z_C \\ 1 \end{bmatrix} = \begin{bmatrix} R & t \\ 0 & 1 \end{bmatrix} \begin{bmatrix} X_W \\ Y_W \\ Z_W \\ 1 \end{bmatrix} \tag{6}$$

further, since:

$$\begin{cases} x = f \cdot X_C / Z_C \\ y = f \cdot Y_C / Z_C \\ u = x/dx + u_0 \\ v = y/dy + v_0 \end{cases} \tag{7}$$

let

$$\begin{cases} f_x = f/dx \\ f_y = f/dy \end{cases} \tag{8}$$

where ($u_0$, $v_0$) are the coordinate of the principal point. *dx* means and *dy* means physical sizes of one pixel in horizontal and vertical direction respectively. The corresponding relationship between pixel and word coordinate can be finally obtained:

$$Z_C \begin{bmatrix} u \\ v \\ 1 \end{bmatrix} = \begin{bmatrix} f_x & 0 & u_0 & 0 \\ 0 & f_y & v_0 & 0 \\ 0 & 0 & 1 & 0 \end{bmatrix} \begin{bmatrix} R & T \\ 0 & 1 \end{bmatrix} \begin{bmatrix} X_w \\ Y_w \\ Z_w \\ 1 \end{bmatrix} \quad (9)$$

Now, we have 3D point cloud, image data and transformation between them, we can generate depth map next. Firstly, point cloud can be projected on to image plane to generate initially sparse depth map. Next, we adopt a self-adaptive method to generate a dense depth map (Chen *et al.*, 2017) according to its excellent performance in keeping consistent with RGB image. Fig. 8 are taken from Chen's paper presents performance of different methods. As we can see from Fig. 8(e), the boundary of dense depth map is well consistent with that of RGB image, which is a vital factor to judge the quality of the depth map generation algorithm. More details about this method can be found in Chen's paper.

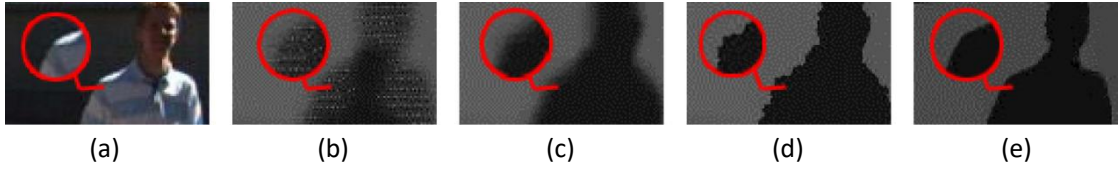

(a)　　　　　(b)　　　　　(c)　　　　　(d)　　　　　(e)

Fig. 8. Visual comparison of depth upsampling result under various conditions. (a) RGB image; (b) dense depth map without depth uncertainty elimination; (c) dense depth map with empirical kernel bandwidth; (d) dense depth map with self-adaptive bandwidth; (e) dense depth map with the global enhancement after self-adaptive bandwidth

Since the generation of depth map depends on the registration parameters which are not accurate and need to be optimized in our paper, the coarse depth map doses help the occlusion detection and assistant the optimization of registration parameters.

**3.2 cost map generation and optimization**

The key idea of the proposed method in our paper is the concept of cost map. Compared with point to point registration, we build a cost map to guide the registration process. Specifically, after obtaining a edge detection result in section 2.2, for each 2D point feature on the image plane, distance transform is used to measure the cost of a 3D point cloud projected on its surrounding area, which means the cost is proportional to the distance to the point feature (Fig. 9(c) reflects this relationship): if the projection location of a 3D point feature is close to a 2D point feature, the corresponding cost is low, otherwise the cost is high. However we don not consider a projection that locates far from a 2D point feature, we set this distance threshold as 50 pixels in our experiments.

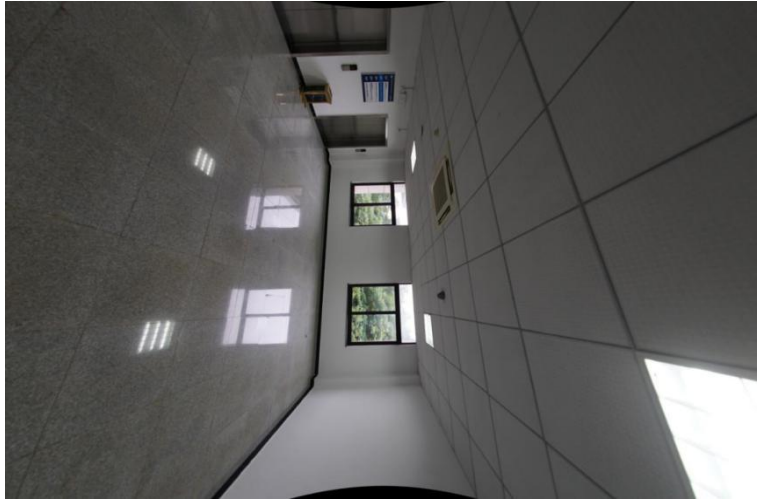
(a) raw RGB image

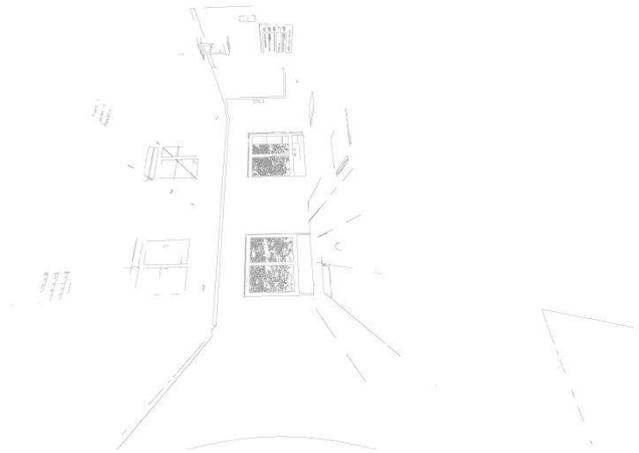
(b) edge detection result

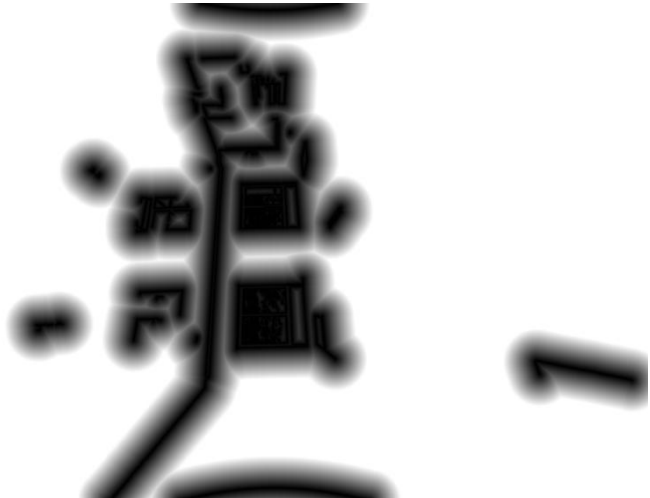
(c) cost map
Fig. 9. Cost map generation

To make it clear for interested readers to repeat our method, some details about optimization need to be emphasized. We use ceres library to realize the whole optimization process. Two things are critical: 1) non-linear least square illustrated on ceres libray is adopted here (ceres solver-b); 2) rotation is transformed to quaternion (ceres solver-a); 3) loss function need to be set to improve the robustness of optimization.

## 4. Experimental Results

We have shown the extraction of 3D point features and 2D point features in section 2.1 and 2.2 respectively, so in this part, we mainly present visual comparisons of initial registration parameters and optimized ones. This may make this part short but clear. Fig 10(a), Fig 11(a) and Fig 12(a) are projection results based on initial registration parameters, and Fig 10(b), Fig 11(b) and Fig 12(b) are projection results in terms of optimized registration parameters. After the optimization, the 3D point features align well with 2D point features.

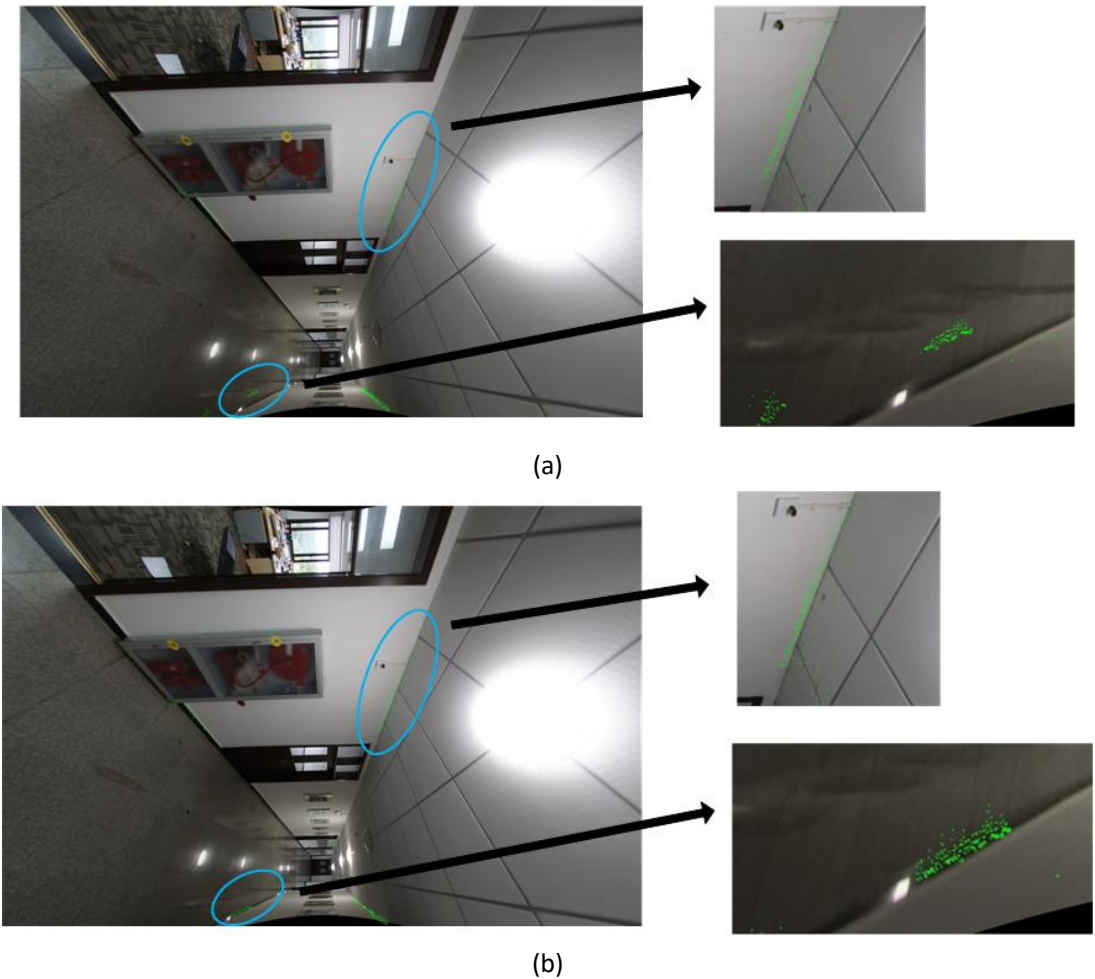

(a)

(b)

Fig. 10. Comparisons without and with registration parameters optimization

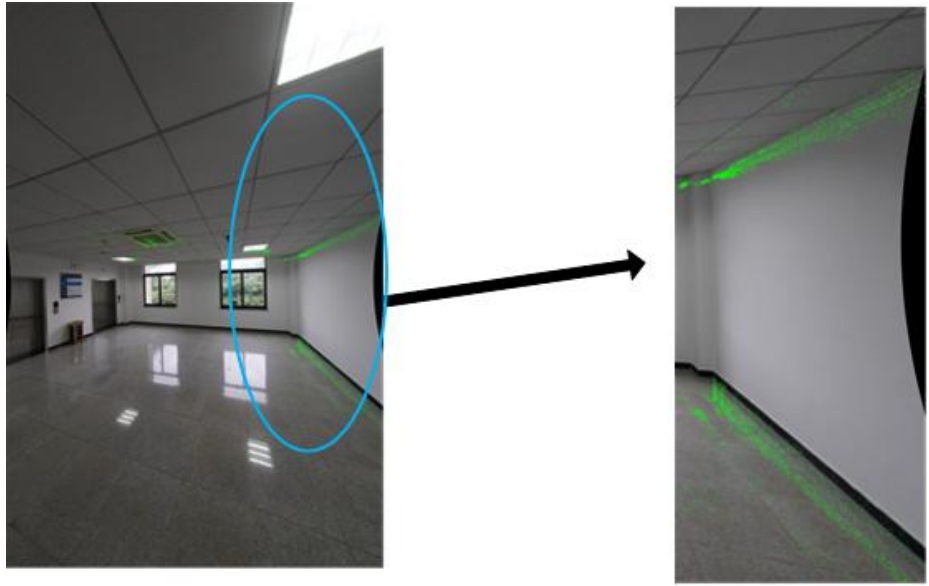

(a)

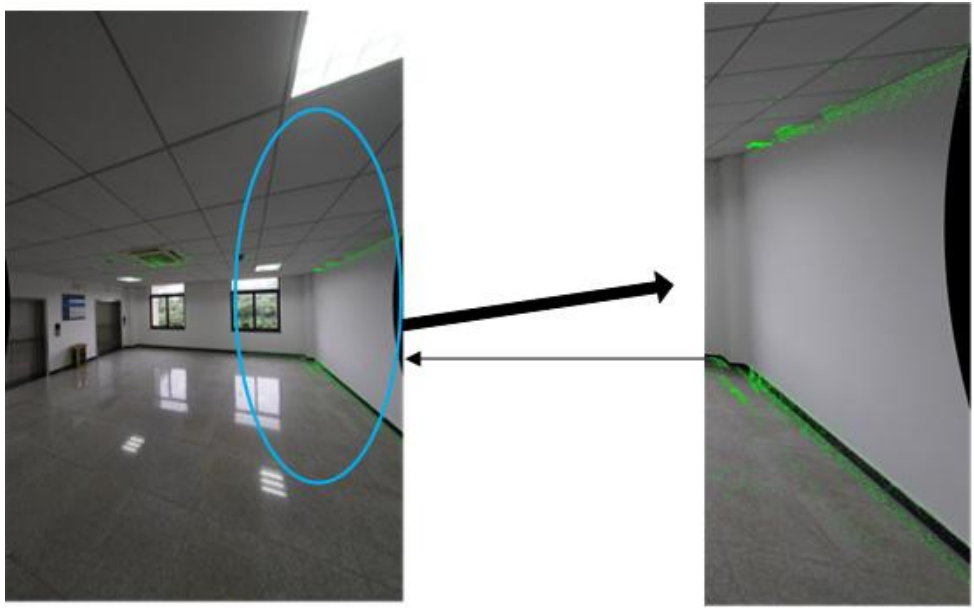

(b)

Fig. 11. Comparisons without and with registration parameters optimization

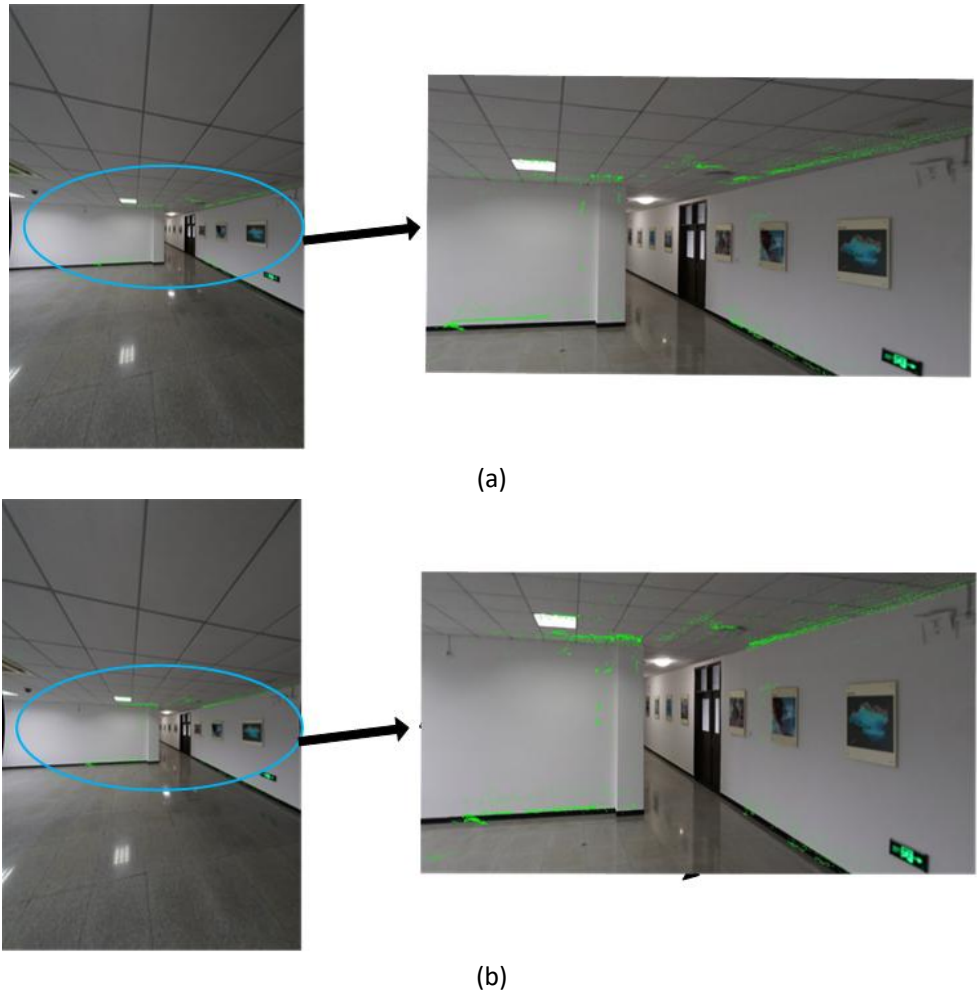

(a)

(b)

Fig. 12. Comparisons without and with registration parameters optimization

## 5. Conclusions

In this paper, we propose a simple and efficient registration of indoor 3D point clouds and optical images based on point features. For 3D point data, point features are extracted from single-frame LiDAR. For image data, classical Canny edge detector is utilized to extract point features. We don't directly search the correspondence features between the two data source, instead, cost map is built to guide the optimization direction. Ceres library is applied as the optimization tool to facilitate the realization of our method. Our method relies on the initial pose data, if the initial error of pose is not heavy, our method should be applicable.

## Acknowledgments

This study was supported in part by the Natural Science Fund of China with Project No. 41874031 and 61872431, by the National Key Research Development Program of China with project No.2016YFB0502204 and 2016YFE0202300, and the Technology Innovation Program of Hubei Province with Project No. 2018AAA070, and the Natural Science Fund of Hubei Province with Project No. 2018CFA007.